\documentclass[letterpaper, 10 pt, conference]{ieeeconf}  % Comment this line out if you need a4paper

\IEEEoverridecommandlockouts                              % This command is only needed if 
\usepackage[noadjust]{cite}

\usepackage{xparse}
\usepackage{amsmath}
\usepackage{amssymb}
\usepackage{amsfonts} 
\usepackage{xcolor}
\usepackage{url}
\usepackage{balance}

\usepackage{graphicx}
\usepackage{subcaption}
\usepackage{etoolbox}

\makeatletter
\let\NAT@parse\undefined
\makeatother
\usepackage[colorlinks]{hyperref}
\usepackage{cleveref}
\Crefname{figure}{Fig.}{Figs.}
\Crefname{equation}{Eq.}{Eqs.}

\usepackage{booktabs}
\usepackage[para]{threeparttable}
\usepackage{multirow}
\usepackage{makecell}
\usepackage{siunitx}
\usepackage{calc}

\title{\LARGE\bf Working Backwards: Learning to Place by Picking}

\author{Oliver Limoyo$^{1,2}$, Abhisek Konar$^{1}$, Trevor Ablett$^{1,2}$, Jonathan Kelly$^{2}$, Francois R. Hogan$^{1}$, and Gregory Dudek$^{1,3}$
\thanks{$^{1}$Authors are with the Samsung AI Centre, Montr\'{e}al, Qu\'{e}bec H3A 3G4, Canada. Email: Email: \{o.limoyo, t.ablett\}@partner.samsung.com, \{abhisek.k,f.hogan,greg.dudek\}@samsung.com}%
\thanks{$^{2}$Authors are with the STARS Laboratory, University of
Toronto Institute for Aerospace Studies, Toronto, Ontario M5S 1A4, Canada. Email: \{\textit{first-name.last-name}\}@robotics.utias.utoronto.ca}%
\thanks{$^{3}$Gregory Dudek is with McGill University, Montr\'{e}al, Qu\'{e}bec H3A 089, Canada. Email: gregory.dudek@mail.mcgill.ca}%
}
\newcommand{\insertfig}{\includegraphics[width=0.90\textwidth]{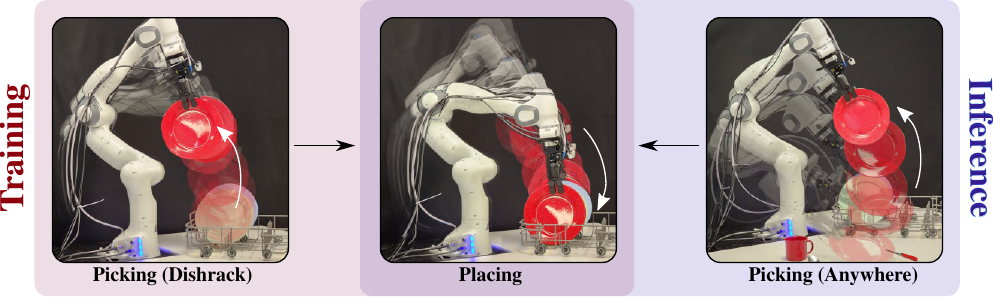}\captionof{figure}{An overview of \textit{placing via picking} (PvP). \textit{Left:} During training, we collect placing demonstrations by reversing the grasping trajectory with objects that are initially at their target locations (in the dish rack). \textit{Right:} During inference, our learned vision-based policy can generalize to object placement scenarios where the objects are not at their target locations initially.} 
\label{fig:title}
\vspace{1mm}
}
\makeatletter
\apptocmd{\@maketitle}{\centering\insertfig}{}{}
\makeatother

\begin{document}
\maketitle

\begin{abstract}
We present \textit{placing via picking} (PvP), a method to autonomously collect real-world demonstrations for a family of placing tasks in which objects must be manipulated to specific, contact-constrained locations.
With PvP, we approach the collection of robotic object placement demonstrations by reversing the grasping process and exploiting the inherent symmetry of the pick and place problems.
Specifically, we obtain placing demonstrations from a set of grasp sequences of objects initially located at their target placement locations. 
Our system can collect hundreds of demonstrations in contact-constrained environments without human intervention using two modules: compliant control for grasping and tactile regrasping.
We train a policy directly from visual observations through behavioural cloning, using the autonomously-collected demonstrations.
By doing so, the policy can generalize to object placement scenarios outside of the training environment without privileged information (e.g., placing a plate picked up from a table).
We validate our approach in home robot scenarios that include dishwasher loading and table setting.
Our approach yields robotic placing policies that outperform policies trained with kinesthetic teaching, both in terms of success rate and data efficiency, while requiring no human supervision.  
\end{abstract}

%%%%%%%%%%%%%%%%%%%%%%%%%%%%%%%%%%%%%%%%%%%%%%%%%%%%%%%%%%%%%%%%%%%%%%%%%%%%%%%%
\section{Introduction} \label{sec:int}

Imitation learning (IL) provides a scaleable, simple, and practical option to learn control policies from expert demonstrations \cite{ablett2021seeing, florence2022implicit}.
The general approach of training expressive models on large and diverse datasets has proven to be quite effective outside of robotics \cite{blattmann2023stable, openai2023gpt4}. 
%
%While there have been many attempts to transfer this paradigm to IL with robotic systems \cite{jang2022bc, lynch2022interactive}, robots have yet to achieve similar successes.  
There have been many attempts to transfer this paradigm to IL for robotic systems \cite{jang2022bc, lynch2022interactive}, but similar successes have yet to be achieved.
A major bottleneck for IL-based approaches in these data regimes is the human time and effort required to collect a large number of expert demonstrations in the physical world.
Having said this, many recent methods attempt to streamline \cite{chi2024universal, zhao23} or automate \cite{kalashnikov2018qt, kalashnikov2021mt, gdm2024autort, bousmalis2023robocat} data collection.
As highlighted by the authors of \cite{gdm2024autort}, the lack of robust and diverse autonomous data collection policies is a major limitation to scaling up IL.
In this work, we are interested in investigating autonomous data collection for robotic object placement.
As opposed to much prior research, we tackle the \emph{full} placing problem, without restricting our task to simple, small objects placed on flat, horizontal surfaces with no environment contact constraints.
We assert that object and environment contact constraints can be a valuable, natural guide for learning how to place, if handled properly.

We propose a novel approach named \textit{placing via picking} (PvP) that automates the collection of expert demonstrations for a large subset of contact-constrained placing problems.
We do so by taking advantage of a powerful grasp planner \cite{sundermeyer2021contact}, tactile sensing, and compliant control.
PvP is a self-supervised pipeline to autonomously collect expert placing demonstrations by leveraging the cyclical nature and inherent symmetry of the pick and place tasks.
Given an environment with objects initially at their target locations, we alternate between picking (i.e., grasping and retrieving) and placing, by time-reversing the retrieval trajectory.
We generate demonstration data for object placement in a self-supervised manner---the picking phase provides the training supervision for the placing task.
While appearing deceptively easy at first, we highlight the crucial importance of two modules, compliant control for grasping (CCG) and tactile regrasping (TR).
CCG and TR enable (1) robust and uninterrupted pick-and-place and (2) the use of environment contact-constraints as guides to inform object placement.
\Cref{fig:title} provides a visual summary of our approach.
Our main contributions are
\begin{enumerate}
\item PvP, a self-supervised data collection method for 6-DOF robotic object placements;
\item a compliant grasping and a tactile-driven regrasping strategy to achieve uninterrupted pick and place data collection in contact-constrained environments;
\item a language-specified grasp planning pipeline; and
\item real-world experimental validation of PvP using collected demonstrations to train a vision-based policy capable of placing multiple plates in a dish rack and of setting a table.
\end{enumerate}

\section{Related Work} \label{sec:rw}

In this section, we review prior work on automated robot data collection, discuss the general idea of ``working backwards,'' and investigate existing robotic placement strategies.

\subsection{Automatic Data Collection}

A large number of existing policy learning approaches use prior policies to automate the data collection process.
Examples include using trajectory optimization \cite{dalal2023imitating, levine2016end} and, closer to our work, simple scripted pick-and-place policies \cite{kalashnikov2018qt, kalashnikov2021mt, gdm2024autort} as forms of supervision.
The authors of \cite{kalashnikov2018qt} use a scripted pick-and-place policy to bootstrap the collection of the initial dataset used for off-policy deep reinforcement learning with a real-world manipulator. 
In \cite{gdm2024autort}, the authors present AutoRT, a method that uses a combination of scripted, learned, and tele-operated data collection policies to efficiently collect a large amount of language-specified manipulation demonstration data. 
Using a combination of a large language model (LLM) and visual language models (VLMs), AutoRT observes scenes in the wild and comes up with plausible tasks.
Once a task is defined and chosen, AutoRT samples a data collection policy to attempt to gather real world data for the task.
In our work, we focus on developing a method for collecting demonstrations of placement with larger and more complex objects (i.e., objects that require a planner to grasp) in environments with contact constraints. 
Unlike prior work, we do not restrict the placing task to simple small objects on flat horizontal surfaces.
The authors of \cite{gdm2024autort} mention the lack of robust and diverse real-world autonomous collection policies as a limitation of AutoRT; our work is complementary to many of the previous frameworks, where PvP could be used as a prior policy.

Other works have explored automatic data generation by bootstrapping from a single human demonstration \cite{johns2021coarse, 10341625}. 
In \cite{johns2021coarse}, a robot manipulation task is modelled as having two phases: a coarse approach trajectory phase towards a bottleneck pose, followed by a fine interaction trajectory phase.
The single human demonstration provides the relative pose of the manipulator with respect to the task-relevant object, from which a self-supervised data collection procedure is formulated to train a bottleneck pose predictor. 
In our work, we also automatically collect and augment approach trajectory data. 
However, we do not require any human demonstrations since we focus on placing tasks only. 
By doing so, we can leverage the inherent symmetry of picking and placing for self-supervised data collection through the use of a combination of tactile sensing, compliant control, and an off-the-shelf grasp planner.

\subsection{Working Backwards}

The general idea of working backwards or time-reversal has been exploited in multiple contexts including generative modelling \cite{sohl2015deep}, learning visual predictive models for control \cite{nair2020trass, limoyo2020heteroscedastic, limoyo2023multimodallatent}, and reinforcement learning \cite{andrychowicz2017hindsight}.
Our self-supervised data collection method, which extends the concept of working backwards to the problem of robotic placement, is most similar in spirit to \cite{fu2022safely, spector2021insertionnet, spector2022insertionnet}. 
These methods consider objects that are assumed to be in their desired place configuration during data collection.
The manipulator can then pick an object and move back to the initial goal location to generate training data.
However, we focus on placement as opposed to insertion.
Insertion policies generally only reason about local interactions. 
On the other hand, placement involves reasoning over a larger scene with the potential for multiple place poses and more complex objects that require planners to grasp successfully.
Furthermore, unlike prior work \cite{fu2022safely, spector2021insertionnet, spector2022insertionnet}, our system does not require a human to guide the manipulator to the object and instead leverages tools from grasp planning.
Our approach also resembles \cite{zakka2020form2fit}, where time-reversed trajectories are used to generate assembly data from disassembly data in a self-supervised manner. 
However, we focus on generating demonstrations for a closed-loop reactive policy that uses a 6-DOF pose-based action representation for placement. 
In contrast, the self-supervised data collection procedure introduced in \cite{zakka2020form2fit} uses a 2D open loop pick-and-place action primitive. 

\subsection{Robotic Manipulator Placing}
Pick and place is a fundamental problem in robotic manipulation. 
A significant amount of research has focused on the picking problem \cite{bicchi2000robotic, bohg2013data, zeng2022robotic}, while placing has been relatively less studied.

Previous approaches have explored the use of motion planning combined with place-specific objectives \cite{haustein2019object, haustein2019placing}. 
For example, in \cite{haustein2019object}, the authors demonstrate a hierarchical sampling-based motion planner that first finds poses which satisfy constraints (e.g., reachability) and then performs a local optimization for user-specified objectives (e.g., clearance). 
The place problem has also been studied from the perspective of geometric modelling \cite{HARADA20141463}, where the planar surfaces of a discretized object model are matched to planar patches in the environment.
Importantly, 3D models of the objects being placed as well as the environment are assumed to be available ahead of time. 
%
%Other works have looked into combining learning with motion planning to operate directly on point clouds, thereby relaxing the assumption of known object and environment models \cite{jiang2012learning, schuster2010perceiving}.

% Another category of methods estimate the pose of the object being grasped via tactile sensing prior to placement or insertion \cite{li2014localization, bauza2022tac2pose}.
% %
% The authors of \cite{li2014localization} localize the pose of a small object using a height map from a tactile sensor and a feature-based matching technique. 
% %
% Given the current in-hand object pose and the place pose, they successfully demonstrate insertions of fine objects. 
% %
% The final place pose and reference tactile maps of the objects are assumed to be known in advance.

%
Closer to our approach, other works have trained end-to-end place policies using reinforcement learning \cite{Dong2021TactileRLFI} and imitation learning \cite{finn2016guided}.
In \cite{finn2016guided}, the authors demonstrate an inverse reinforcement learning technique capable of simultaneously learning a cost function and a control policy from human kinesthetic demonstrations of dish placing.    
The main contribution and novelty of our work is a data collection technique for robotic placing tasks that can be used in tandem with methods that require expert demonstrations.

\section{Placing via Picking} \label{sec:me}

We first present an overview of our self-supervised data collection method, PvP, in \Cref{subsec:me_data}, and then describe a method for noise augmentation during data collection in \Cref{subsec:noiseaugment}.
Finally, we provide details about the procedure used to train our IL policy in \Cref{subsec:policy}. 

\subsection{Self-Supervised Data Collection} \label{subsec:me_data}
\begin{figure*}
  \setlength{\fboxsep}{0pt}%
  \setlength{\fboxrule}{1pt}%
  \centering
  \begin{subfigure}{0.245\textwidth}
    \centering
    \fbox{\includegraphics[height=1.6in-2pt]{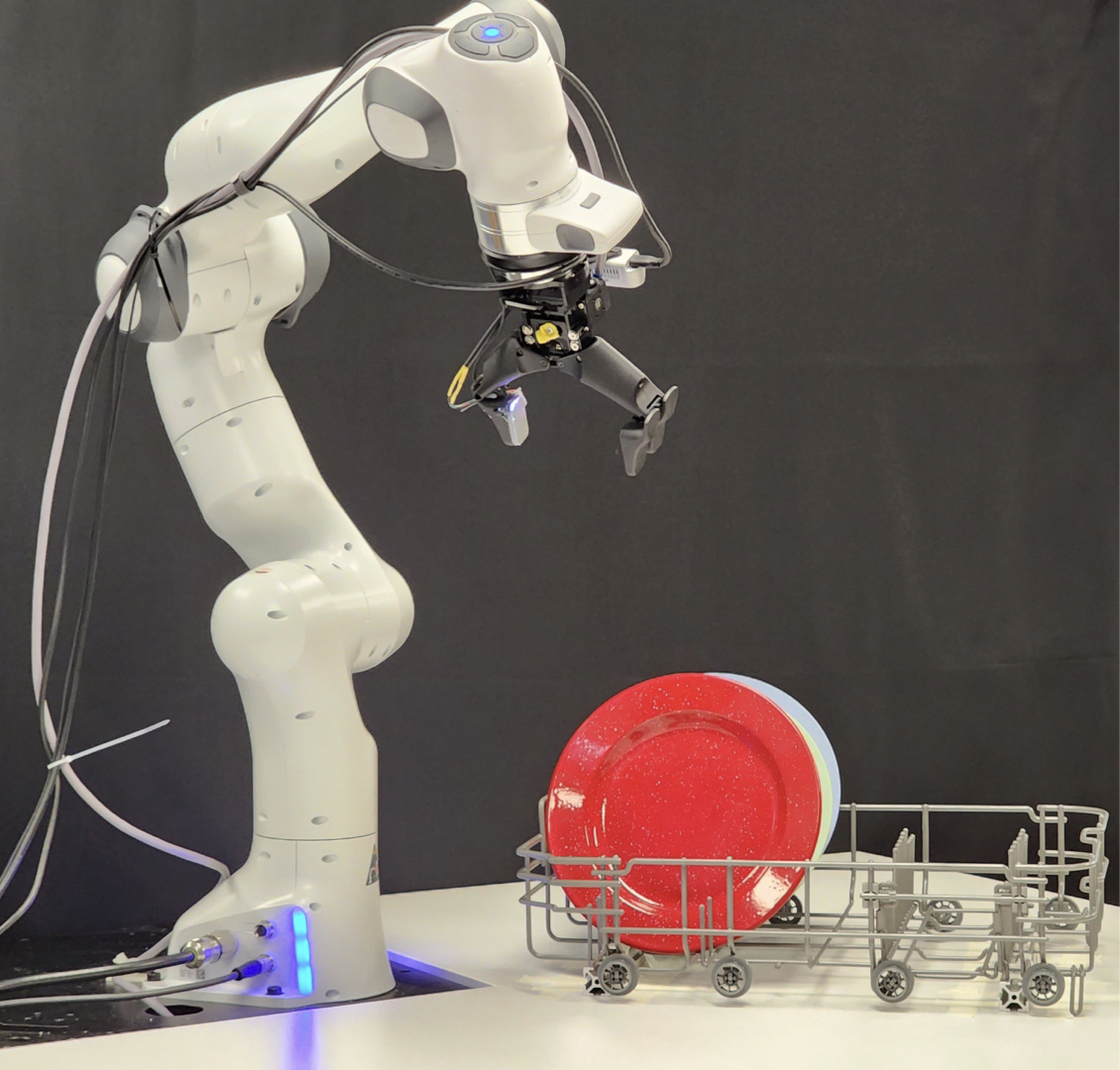}}
    \subcaption{Grasp planning.}
    \label{subfig:scan}
  \end{subfigure}
  \hfill
  \begin{subfigure}{0.245\textwidth}
    \centering
    \fbox{\includegraphics[height=1.6in-2pt]{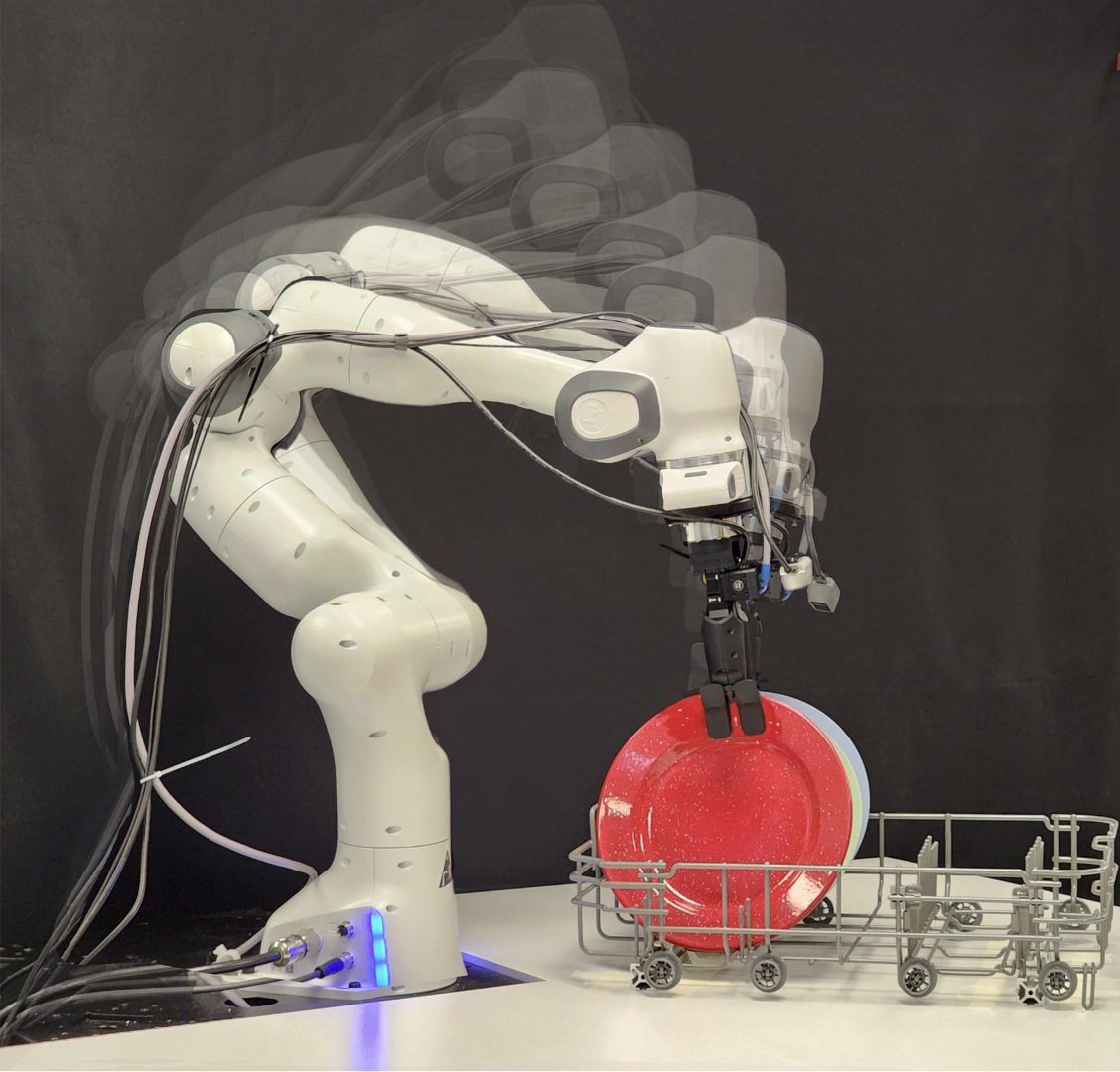}}
    \subcaption{Grasping.}
    \label{subfig:grasp}
  \end{subfigure}
  \hfill
  \begin{subfigure}{0.245\textwidth}
    \centering
    \fbox{\includegraphics[height=1.6in-2pt]{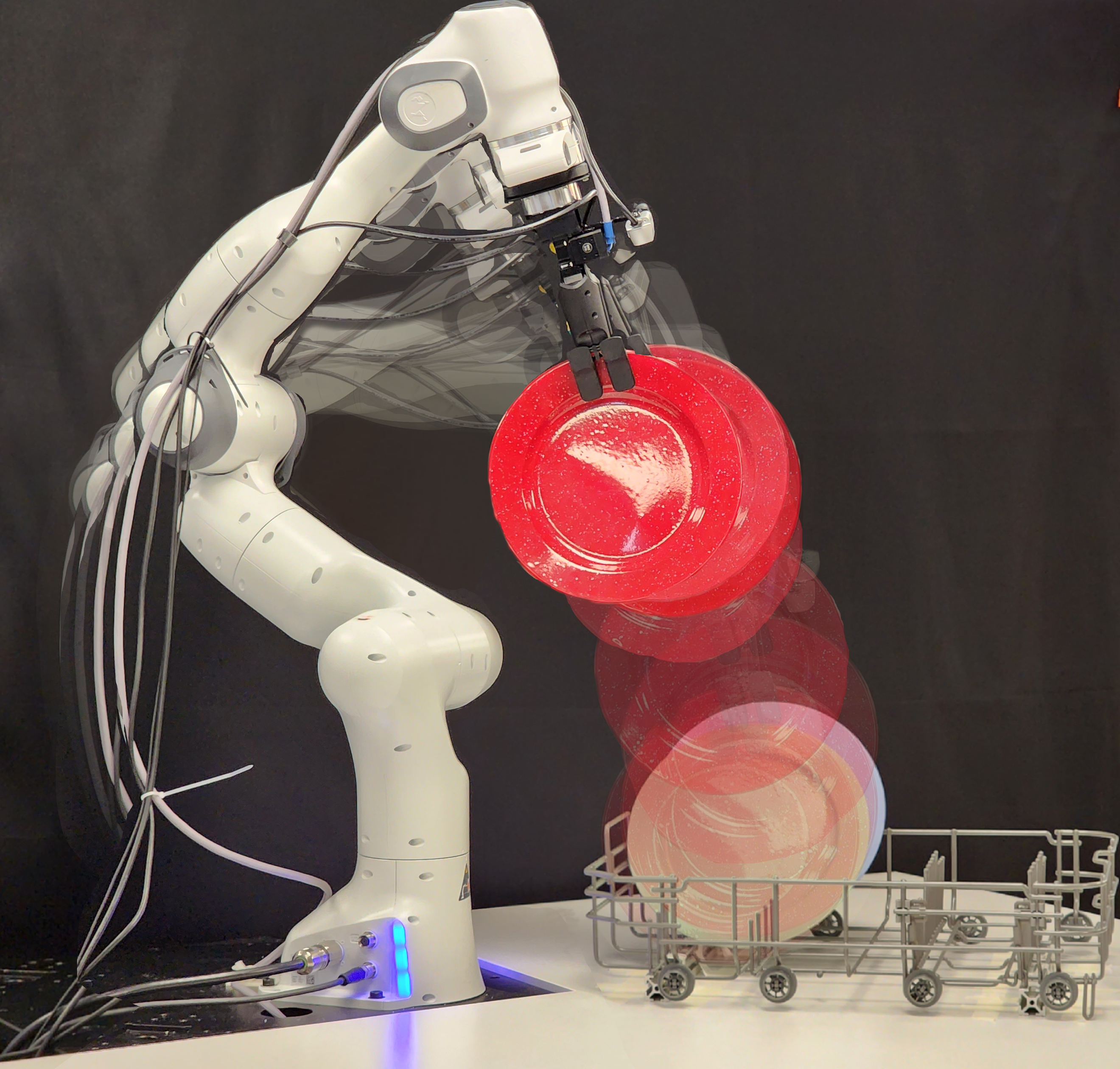}}
    \subcaption{Retrieving.}
    \label{subfig:retrieve}
  \end{subfigure}
  \hfill
  \begin{subfigure}{0.245\textwidth}
    \centering
	\fbox{\includegraphics[height=1.6in-2pt]{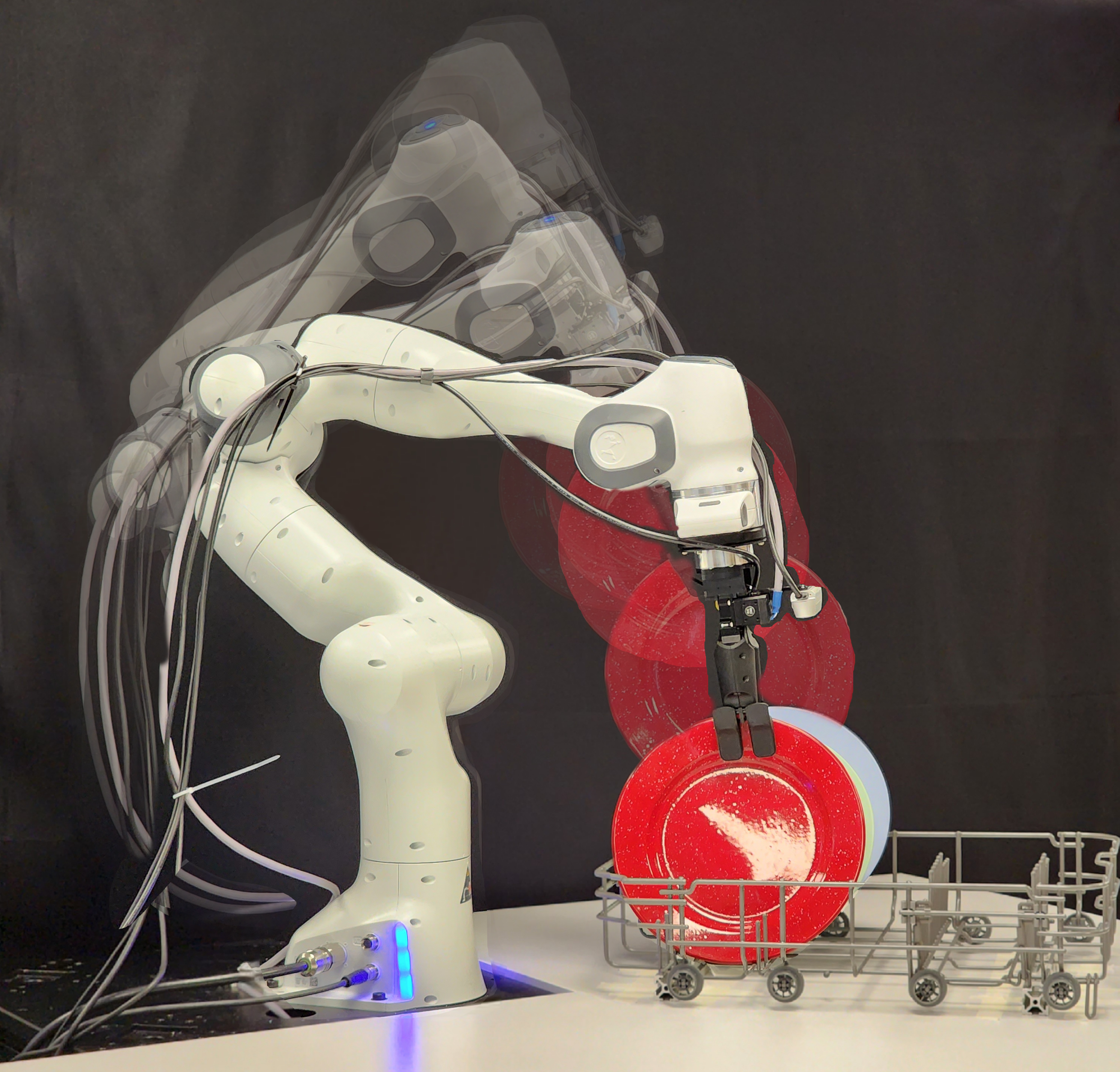}}
    \subcaption{Placing by reversing.}
    \label{subfig:place}
  \end{subfigure}
  \caption{The four steps involved in PvP, our autonomous demonstration data collection process for placing. We (a) generate grasps with an off-the-shelf grasp planner \cite{sundermeyer2021contact}; (b) compliantly grasp the object to apply minimal forces to the environment while ensuring a stable grasp via tactile sensing; (c) retrieve the object with rotational compliance while storing the trajectory; and (d) generate placement demonstration data by rolling out the reversed grasp trajectories while storing the observations and actions.}
  \label{fig:methodoverview}
%\vspace{-1mm}
\end{figure*}

PvP consists of a cycle with four main phases: (1) grasp planning, (2) grasping, (3) retrieving, and (4) placing by reversing; these steps are visualized in \Cref{fig:methodoverview}. 
During training, the objects of interest are initially in their goal positions.

\subsubsection{Grasp Planning}
\begin{figure*}
  \centering
  \setlength{\fboxsep}{0pt}%
  \setlength{\fboxrule}{1pt}%
  \begin{subfigure}{0.245\textwidth}
    \centering
    \fbox{\includegraphics[height=1.252in-2pt]{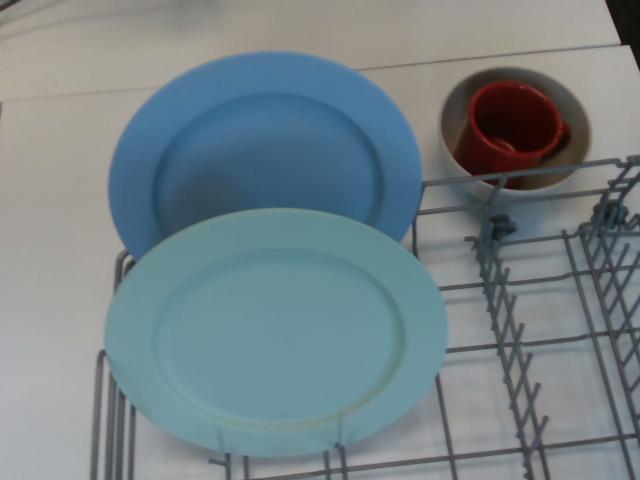}} \\
    \vspace{1mm}
    \fbox{\includegraphics[height=1.252in-2pt]{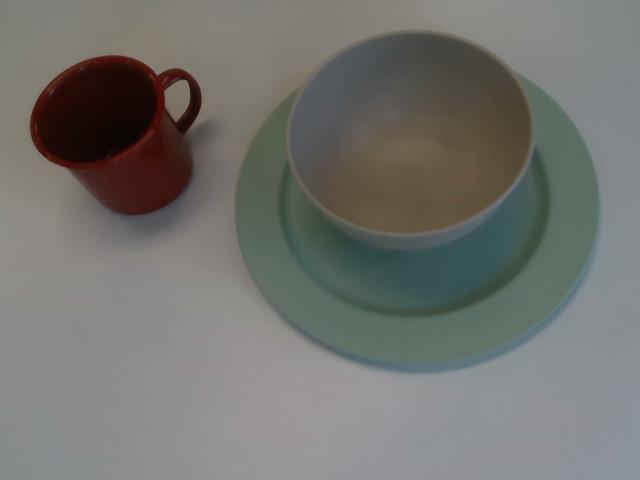}}
    \subcaption{Raw image.}
    \label{subfig:raw}
  \end{subfigure}
  \begin{subfigure}{0.245\textwidth}
    \centering
    \fbox{\includegraphics[height=1.252in-2pt]{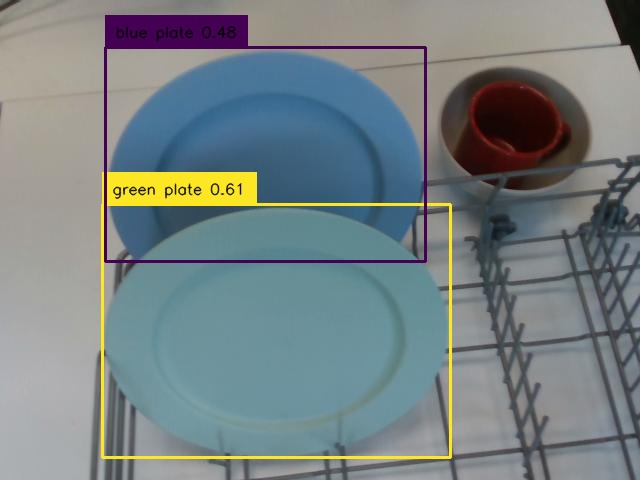}} \\
    \vspace{1mm}
    \fbox{\includegraphics[height=1.252in-2pt]{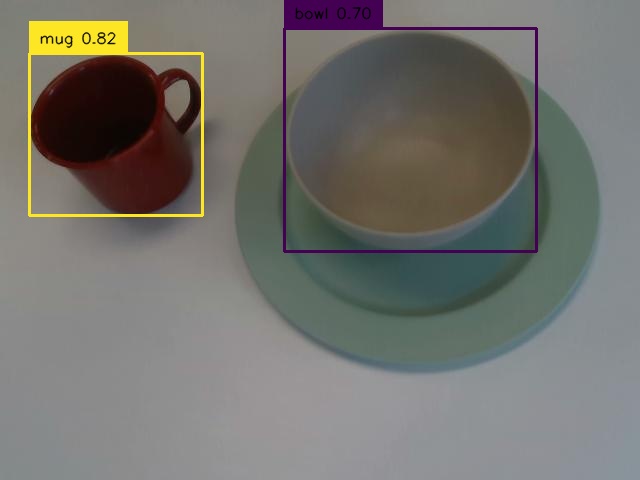}}
    \subcaption{Text-based object detection.}
    \label{subfig:detect}
  \end{subfigure}
  \begin{subfigure}{0.245\textwidth}
    \centering
    \fbox{\includegraphics[height=1.252in-2pt]{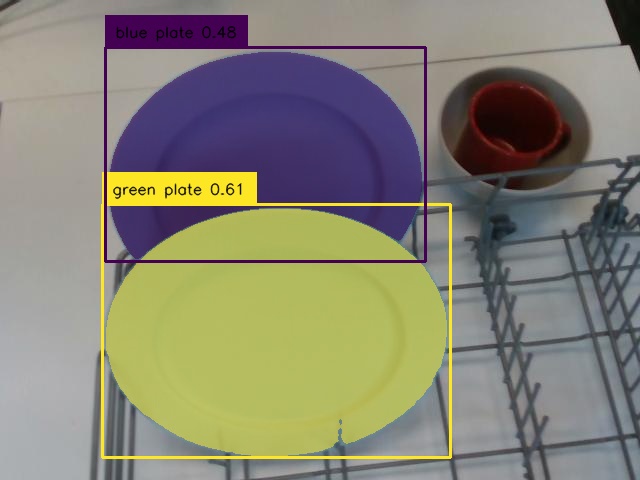}} \\
    \vspace{1mm}
    \fbox{\includegraphics[height=1.252in-2pt]{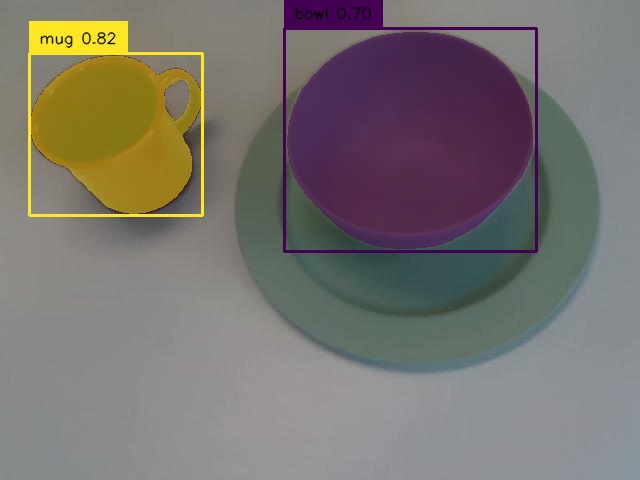}}
    \subcaption{Segmentation.}
    \label{subfig:seg}
  \end{subfigure}
  \begin{subfigure}{0.245\textwidth}
    \centering
    \fbox{\includegraphics[height=1.252in-2pt]{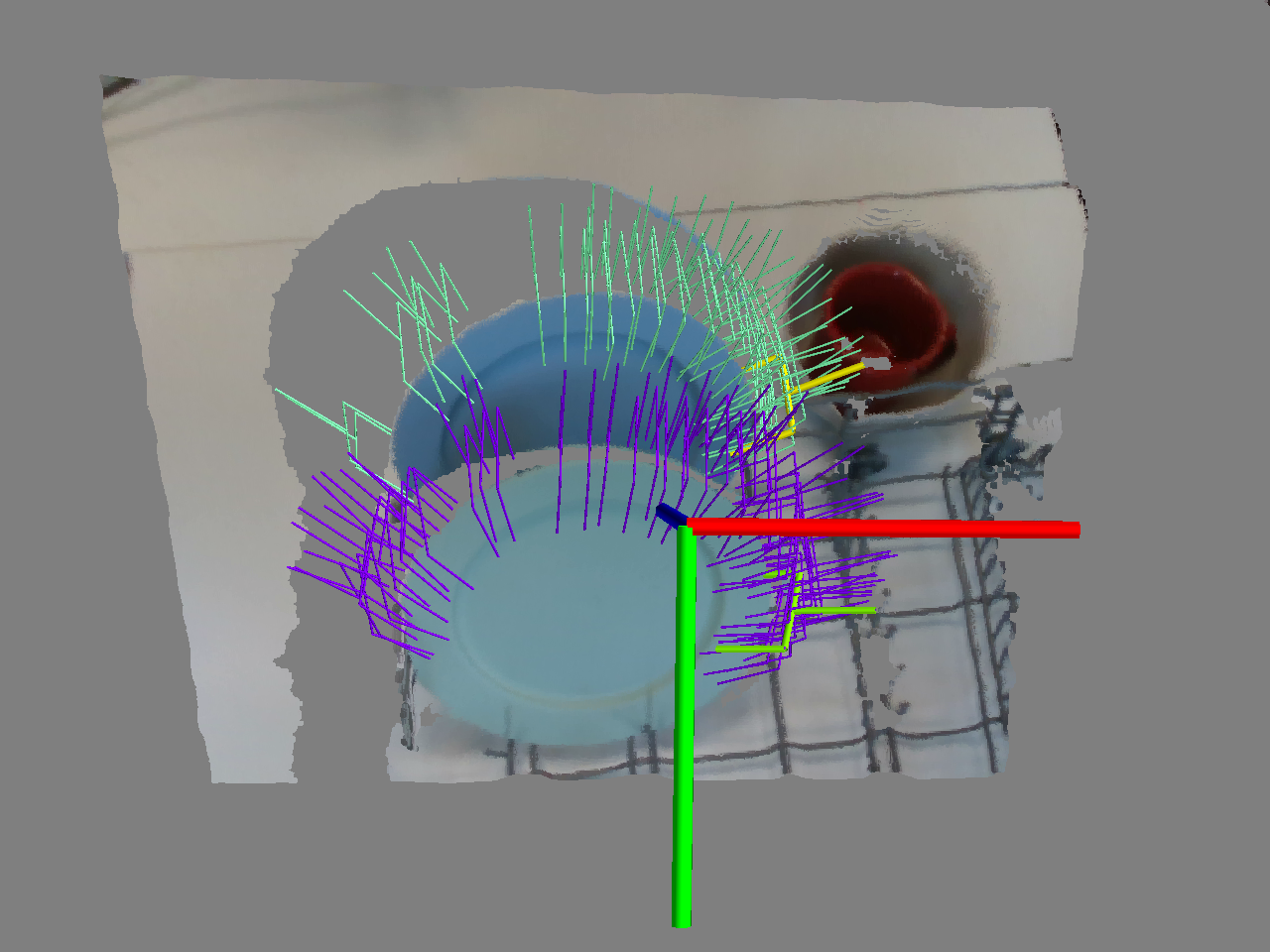}} \\
    \vspace{1mm}
    \fbox{\includegraphics[height=1.252in-2pt]{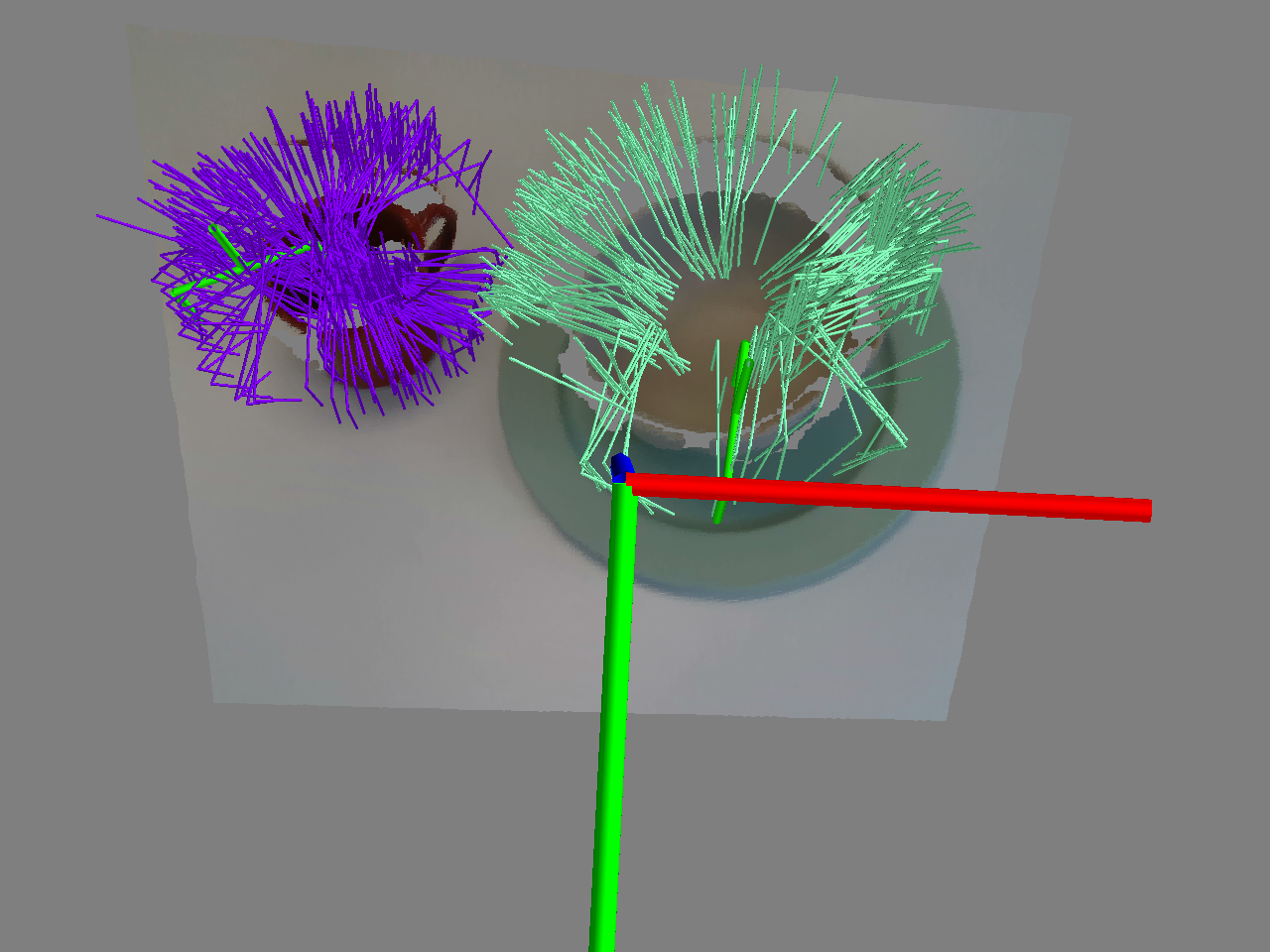}}
    \subcaption{Pruned grasps.}
    \label{subfig:grasps}
  \end{subfigure}
  \caption{The steps involved in our language-driven grasp planning pipeline. We use (b) Grounding-Dino \cite{liu2023grounding} for object bounding box detection based on text descriptions and (c) Segment Anything \cite{kirillov2023segany} on the cropped images to produce object specific masks. We then use (d) Contact-GraspNet \cite{sundermeyer2021contact} for grasp generation on only the segmented objects (i.e., masked areas). \textit{Top}: using ``green plate" and ``blue plate'' as the targets for data collection. \textit{Bottom}: using ``bowl'' and ``mug'' as the targets for data collection. }
  \label{fig:graspoverview}
\vspace{-4mm}
\end{figure*}

At the start of the data collection process, the manipulator moves to a pre-determined pose where the camera has a clear view of the objects of interest in the environment, as shown in \Cref{subfig:scan}. 
We use the grasp planner Contact-GraspNet \cite{sundermeyer2021contact} to generate $L$ candidate grasp poses $\{\mathbf{T}_{g,i}\}_{i=0}^{L}$, where $\mathbf{T}_{g,i} \in \mathrm{SE}(3)$ for all $i \in \{0, ..., L\}$.
Given a set of text-based descriptions of the objects of interest (e.g., ``plates,'' ``green plate,'' ``cup'' or, more generally, ``objects''), potentially from a user or high-level LLM planner, we use Grounding-Dino \cite{liu2023grounding} to find the object bounding boxes.
Finally, we run Segment Anything \cite{kirillov2023segany} on each of the previously-generated image crops from the bounding boxes to generate per-object masks.
Contact-GraspNet then uses these masks to filter grasps that are not on objects of interest and keeps only a pruned set of grasps $\{\mathbf{T}_{g,i}\}_{i=0}^{K}$, where $K \le L$.
We visualize an example of this process in \Cref{fig:graspoverview} for two different scenes.
After pruning, we randomly select a single grasp pose, $\mathbf{T}_{\text{grasp}}$, from $\{\mathbf{T}_{g,i}\}_{i=0}^{K}$.
To approach without collisions, we include a pregrasp pose $\mathbf{T}_{\text{pregrasp}}$ that is defined with a translational offset above the grasp pose.

\subsubsection{Grasping} \label{subsubsec:grasping}
Given the computed pregrasp and grasp poses, we move to each pose sequentially by linearly and spherically linearly interpolating to produce a smooth trajectory.
An example of a grasp sequence is shown in \Cref{subfig:grasp}.
During grasping, two modules play a critical role in the robustness of PvP: compliant control for grasping (CCG) and tactile regrasping (TR).

We use a Cartesian impedance controller for manipulation.
This choice allows us to set the level of compliance of the manipulator at various stages as needed. 
Before closing the gripper to grasp, we set the translational and rotational stiffness of the manipulator's controller to minimal values. 
We call this CCG.
The high compliance of the robot arm allows the manipulator and gripper to comply with the contact constraints and minimize the applied forces against the environment.
Larger contact-constrained objects generate unnecessary reaction forces when gripped without compliance.
Large forces, in turn, trigger emergency stops and can damage the environment and the robot.
Specific to PvP, large reaction forces also disrupt and move the object relative to the gripper, ultimately leading to inaccurate placements.
On the other hand, when grasping with compliance, the environment contact constraints guide the manipulator towards a natural object placement pose.

\begin{figure}[b]
  \vspace{-2mm}
  \centering
  \includegraphics[width=0.99\columnwidth]{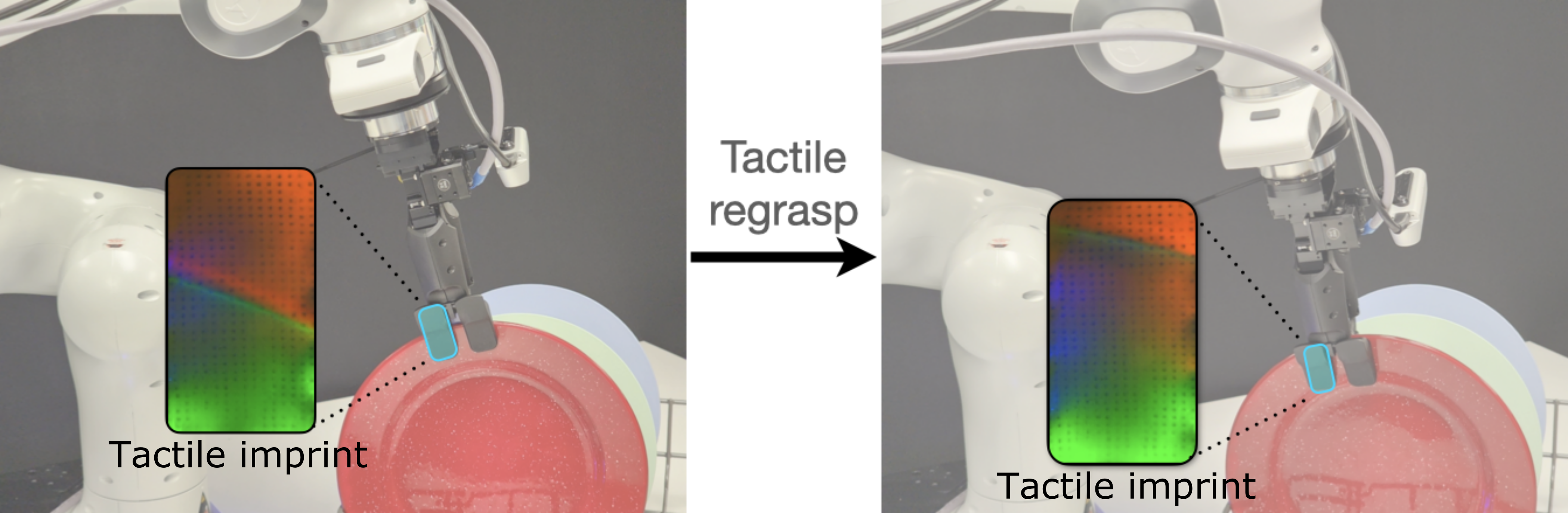}
  \caption{Tactile images from before and after a tactile regrasp (TR). Left: the contact surface area of the plate fills half of the tactile image, indicating a shallow grasp. Right: after a regrasp, the contact surface area has increased, indicating a stable grasp.}
  \label{fig:tactileregrasp}
\end{figure}

\begin{figure}[t]
  \vspace{2mm}
  \centering
  \begin{subfigure}{0.24\columnwidth}
    \includegraphics[height=6cm, keepaspectratio]{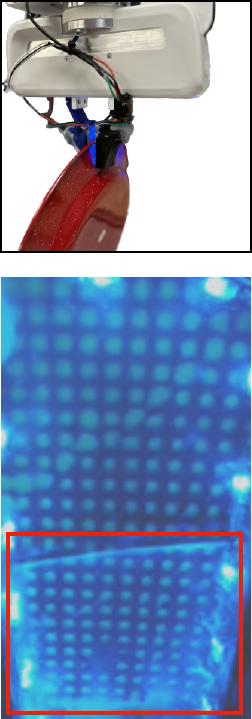}
    \subcaption{}
    \label{subfig:edge1}
  \end{subfigure}
  \begin{subfigure}{0.24\columnwidth}
    \includegraphics[height=6cm, keepaspectratio]{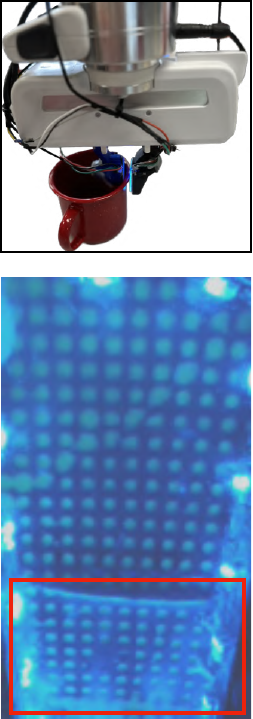}
    \subcaption{}
    \label{subfig:edge2}
  \end{subfigure}
  \begin{subfigure}{0.24\columnwidth}
    \includegraphics[height=6cm, keepaspectratio]{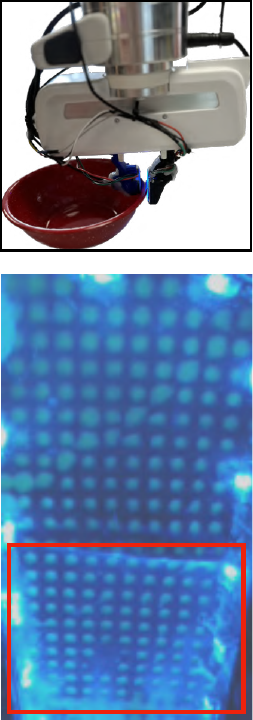}
    \subcaption{}
    \label{subfig:edge3}
  \end{subfigure}
  \begin{subfigure}{0.24\columnwidth}
    \includegraphics[height=6cm, keepaspectratio]{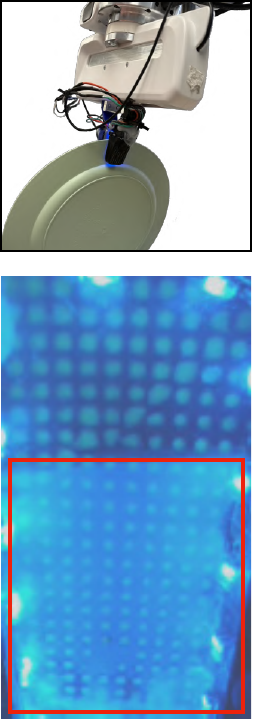}
    \subcaption{}
    \label{subfig:edge4}
  \end{subfigure}
  \caption{Visualization of contact surface region in the tactile image of grasps of various objects: (a) a red metal plate, (b) a red metal mug, (c) a red metal bowl and (d) a green wheat straw plate.}
  \label{fig:generaledge}
  \vspace{-2mm}
\end{figure}

Our method assumes that the grasps that we perform are \emph{stable}, that is, the object does not move or slip significantly once grasped.
In the real world, the accumulation of errors from various sources (e.g., camera calibration, controller, learned grasp planner, and noise from the RGB-D sensor and manipulator encoders) can result in grasps that are unstable (e.g., shallow grasps) and not ideal for PvP.
To mitigate this issue, we use tactile sensing---specifically, a visuotactile Finger-STS \cite{9832483} sensor---to preemptively detect if a grasp is stable and perform a regrasp if needed.
The Finger-STS is capable of providing multimodal feedback (i.e., both tactile and visual observations).
To this end, we design our TR module to detect the contact surface area and calculate the relative change in end effector (EE) pose required to recover a larger contact surface and, thereby, a more stable grasp. 
We detect regions of contact based on marker displacements and the RGB image from the Finger-STS.
If partial contact is detected, we calculate the difference between the ideal and detected contact region and use this to calculate the commanded change in EE pose.
We note that a simpler and more general random regrasp strategy could also be used.
An example of this is shown in \Cref{fig:tactileregrasp}.
We visualize contact regions for various household items in \Cref{fig:generaledge}.

In \Cref{subsec:robust}, we experimentally investigate the effects of including these modules through an ablation study.

\subsubsection{Retrieving}
We begin the retrieval phase when a valid, stable grasp is detected.
An image of the retrieval procedure is shown in \Cref{subfig:retrieve}.
We store the poses of the EE as well as the respective timestamps throughout the retrieval process.
For our work, we measure joint encoder readings at a rate of 120 \si{Hz}.
The manipulator is set to be compliant along the rotational axes of the EE only, when commanded to move to the pregrasp pose. 
%
% This allows the manipulator to better comply with any unexpected environment interactions. 
%
Once at the pregrasp pose, we randomly sample a clearance pose $\mathbf{T}_{\text{clearance}}$ around a set fixed pose above the scene.
That is, given a set fixed pose, we add random translations sampled from $\mathcal{N}(0,\,\sigma_{tr}^{2})$ to recover a clearance pose.
We use $\sigma_{tr}$ = 2.5 cm during data collection.

At the end of the retrieval phase, we have an expert retrieval trajectory of length $M$ from grasp pose to clearance pose via a pregrasp pose,
\begin{equation}
    \tau^{\text{expert}}_{r} = ((\mathbf{T}_{0}, t_{0}), ..., (\mathbf{T}_{M}, t_{M})),
\end{equation}
where $\mathbf{T}_{0} = \mathbf{T}_{\text{grasp}}$, $\mathbf{T}_{M} = \mathbf{T}_{\text{clearance}}$ and $t_{0}, ..., t_{M}$ are the respective timestamps of the trajectory starting with $t_{0}$ = 0.
To match the desired control frequency of our policy, we sample states $\Delta t$ apart to generate a sparser trajectory $\bar{\tau}^{\text{expert}}_{r}$.
 We extract a total of $\bar{M} + 1$ states, where $\bar{M} = \lceil \frac{t_{M} - t_{0}}{\Delta t} \rceil$, by finding the nearest pose (in terms of time) from the dense trajectory $\tau^{\text{expert}}_{r}$ at every $\Delta t$ interval:
\begin{equation}
    \bar{\tau}^{\text{expert}}_{r} = ((\bar{\mathbf{T}}_{0}, 0), (\bar{\mathbf{T}}_{1}, \Delta t), ..., (\bar{\mathbf{T}}_{\bar{M}}, \bar{M}\Delta t)).
\end{equation}
In this work, our visual policies operate at 5 \si{Hz}, so we set $\Delta t =$ 0.20 s.

\subsubsection{Placing by Reversing}

We reverse the sparse retrieval trajectory $\bar{\tau}^{\text{expert}}_{r}$ to extract a desired place trajectory, 
\begin{equation}
    \bar{\tau}^{\text{expert}}_{p} = (\bar{\mathbf{T}}_{\bar{M}}, ..., \bar{\mathbf{T}}_{0}).
\end{equation}
An example of a place trajectory is shown in \Cref{subfig:place}.
Additionally, we convert the global poses to relative poses in order to use state differences as expert actions:
\begin{equation}
    \Delta\mathbf{T}_{i} = \bar{\mathbf{T}}_{\bar{M}-i}^{-1}\bar{\mathbf{T}}^{\phantom{-1}}_{\bar{M}-i-1},
\end{equation}
where $i \in \{0, ..., \bar{M} - 1\}$ and $\Delta\mathbf{T}_{i} \in \mathrm{SE}(3)$ is a relative change in EE pose.
We define the expert action as $\mathbf{a}_{i} = (\Delta\mathbf{T}_{i}, a_{\text{gripper}, i})$, where $a_{\text{gripper}, i} \in \{0, 1\}$ is a binary variable describing the gripper command (open or close). 
To collect an episode of expert data, we perform a rollout of the downsampled place commands, $\{\mathbf{a}_{i}\}_{i=0}^{\bar{M} - 1}$, and store the sequences of observations and actions as an expert trajectory.

We use RGB images from the wrist camera to form the robot's observation space $\mathbf{o}_{i}$. 
We downsample the images to a resolution of 128 $\times$ 128 and also stack the three previous RGB frames, which results in $\mathbf{o}_{i} \in \mathbb{R}^{128 \times 128 \times 12}$.
During the place demonstration, the robot keeps its gripper closed (i.e., the robot uses the gripper command $a_{\text{gripper}, i} = 1$ for all $i \in \{0, ..., \bar{M}-1\}$).
Finally, at the end of the trajectory, we command the robot to keep the EE in its current pose and to open the gripper (i.e. $\Delta\mathbf{T}_{i} = \boldsymbol{\mathbf{I}}$ and $a_{\text{gripper}, i} = 0$) for $N$ time steps.
We use $N=$ 5, which gives the robot an adequate amount of time to open the gripper and release the object.
A single expert demonstration or trajectory is then defined as a set of training tuples
\begin{equation}
    \tau^{\text{expert}} = \{\mathbf{(o}_{t}, \mathbf{a}_{t}, \mathbf{o}_{t+1})\}_{t=0}^{T},
\end{equation}
with a total length of $T=\bar{M}+N$.

\subsection{Noise-Augmented Data Collection} \label{subsec:noiseaugment}

To increase the coverage of our demonstrations and train more robust policies, we perturb the first 75\% of the poses with noise.
We take inspiration from previous works that have investigated the use of noise injection for both the policy observations \cite{florence2019self}, and actions \cite{laskey2017dart}.
We also take inspiration from robust control theory, where injecting isotropic Gaussian noise has been shown to achieve \textit{persistent excitation} \cite{GREEN1986351}, a condition where the training data is informative enough to learn a model that is robust to compounding errors during deployment. 
We call this variant \textit{noise-augmented data collection}. 
In practice, we represent pose commands $\Delta\mathbf{T}$ as a combination of a translation vector $\mathbf{t} \in \mathbb{R}^{3}$ and a rotation vector $\boldsymbol{\theta} \in \mathbb{R}^{3}$.
The rotation vector is defined as $\boldsymbol{\theta} = \theta \mathbf{e}$, where $\theta \in \mathbb{R}$ is the rotation angle and $\mathbf{e} \in \mathbb{R}^{3}$ is the rotation axis. 
We perturb the translation of the pose with
\begin{equation}
    \mathbf{t}_{\text{perturbed}} = \mathbf{t} + \delta \mathbf{t},
\end{equation}
where $\delta \mathbf{t} \in \mathbb{R}^{3}$ is an isotropic Gaussian noise vector and each dimension of $\delta \mathbf{t}$ is sampled from $\mathcal{N}(0,\sigma_{t})$.
Similarly, we perturb the rotation with
\begin{equation}
    \boldsymbol{\theta}_{\text{perturbed}} = (\theta + \delta \theta) (\mathbf{e} + \delta \mathbf{e}),
\end{equation}
where $\delta \mathbf{e} \in \mathbb{R}^{3}$ is an isotropic Gaussian noise vector and each dimension of $\delta \mathbf{e}$ is sampled from $\mathcal{N}(0,\sigma_{e})$, and $\delta \theta \in \mathbb{R}$ is sampled from $\mathcal{N}(0,\sigma_{\theta})$.
We perturb both the rotation axis and the rotation angle.
We find $\sigma_{t}=$ 0.5 cm, $\sigma_{e}=$ 0.5 cm and $\sigma_{\theta}=$ 0.5$^{\circ}$ to be reasonable values.
We evaluate the effects of this augmentation on policy performance in \Cref{subsec:noise}
\subsection{Policy Learning} \label{subsec:policy}
We learn a policy $\pi_{\boldsymbol{\phi}}(\mathbf{a}_{t} \mid \mathbf{o}_{t})$ parameterized by $\boldsymbol{\phi}$ using data generated by PvP.
We use behavioural cloning (BC) \cite{Pomerleau-1989-15721} to train our policy with a likelihood-based loss or objective function.
Given a dataset of $N_{train}$ demonstration trajectories, our loss function is then:
\begin{equation}
    \mathcal{L} = \frac{1}{N_{\text{train}}} \sum_{i=1}^{N_{\text{train}}} \sum_{t=0}^{T} -\log{\pi_{\phi}(\mathbf{a}_{t} \mid \mathbf{o}_{t})}.
\end{equation}
The main backbone of our policy network consists of a convolutional neural network (CNN) based on the ResNet18 architecture \cite{he2016deep}, followed by a multilayer perceptron (MLP) layer that maps the CNN output into the parameters of the distribution of actions.
In this work, we consider two representations for the action distribution $\pi_{\phi}(\mathbf{a}\!\! \mid\!\! \mathbf{o})$: a unimodal Gaussian and a multimodal mixture of Gaussians.
In both cases, during inference or control, we use a low-noise evaluation scheme, as done in \cite{wang2020critic} and \cite{Mandlekar2021WhatMI}, by setting the standard deviation of the Gaussian distributions to be an arbitrarily small value. 
Note that the unimodal Gaussian is then equivalent to a standard deterministic policy trained with a mean squared error (MSE) loss. 
%
% We use a learning rate of $1e-4$ with AdamW \cite{LoshchilovH19} as our optimizer with a weight decay rate of $0.1$, which is only applied to the linear and convolutional layers.

\section{Experiments} \label{sec:exp}

\begin{figure*}[]
  \centering
  \begin{subfigure}{\textwidth}
    \centering
    \includegraphics[width=\textwidth]{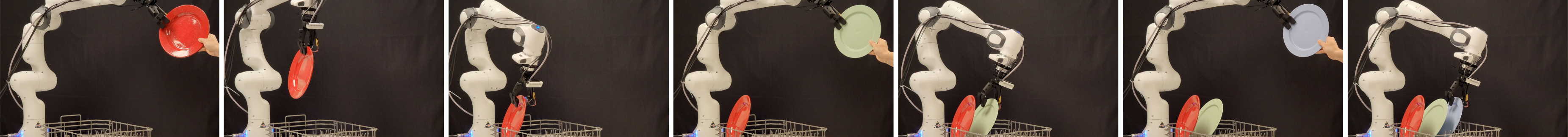}
    \subcaption{Dishrack placement task, which consists of placing multiple plates of varying physical properties in evenly spaced slots.}
    \label{subfig:plateplace}
  \end{subfigure}
  \par\medskip
  \begin{subfigure}{\textwidth}
    \centering    
    \includegraphics[width=\textwidth]{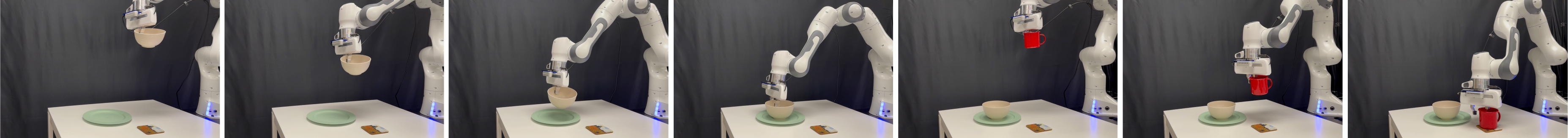}
    \subcaption{Table placement task, which consists of placing a bowl on a plate and a cup on a coaster.}
    \label{subfig:tableplace}
  \end{subfigure}
  \caption{Sequence of images from roll outs of place policies trained using data collected with PvP. The policies are able to place objects of varying properties in the scene using images from the wrist camera directly.}
  \label{fig:rollouts}
\vspace{-2mm}
\end{figure*}

We used PvP to collect data and train robotic placement policies for two separate tasks: dishrask loading and table setting, as shown in \Cref{fig:rollouts}.
In both cases, the place policy had to account for the weight and shape of each item, the initial grasp of the object, and the state of the current scene to decide where and how to place (e.g., place angle). 
In this section, we only present our quantitative results for the more challenging dishrack loading task as it better represents the type of task (with added contact constraints from the environment) that PvP was designed for.
We study the robustness of PvP in \Cref{subsec:robust}, the effect of noise augmentation on the performance of the learned placing policy in \Cref{subsec:noise}, and the relative quality of policies trained with data from PvP and from kinesthetic teaching in \Cref{subsec:comparison}. 

\subsection{PvP Data Collection Robustness} \label{subsec:robust}

We require a robust data collection loop to collect a large number of place demonstrations autonomously with PvP.
We studied the effects of CCG and TR, as introduced in \Cref{sec:me}, on the robustness of PvP.

We observed two main causes of misplaced objects during data collection, initially discussed in \Cref{subsubsec:grasping},
First, the manipulator would stiffly grasp the object causing it to move and generate a large amount of force preload against the environment.
During retrieval, significant object motion occurred at the moment when the object's contact force with the environment was released.
This unexpected shift in the object pose lead to less accurate placements since the reversal procedure no longer resulted in correct alignment between the object and the environment.
With PvP, we mitigate this failure mode with CCG.
The second failure mode involved unstable grasps that lead to a large relative motion between the object and the gripper. 
With PvP, we mitigate unstable grasps using TR to preemptively detect unstable grasps and to regrasp if needed. 

\begin{figure}[b]
  \vspace{-2mm}
  \centering
  \includegraphics[width=0.75\columnwidth]{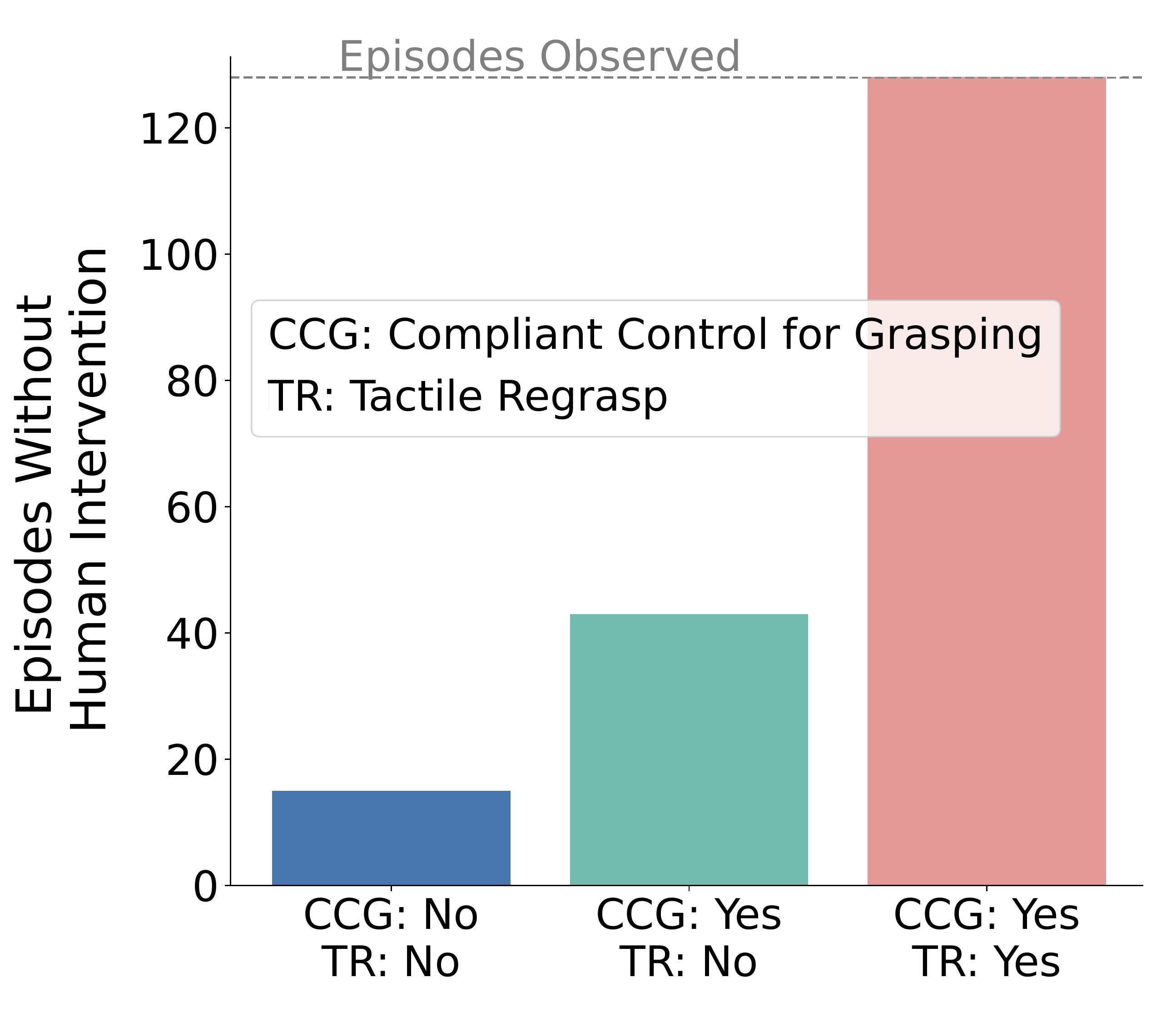}
  \caption{Ablation study on compliant control for grasping (CCG) and tactile regrasping (TR). We measured the average number of episodes that could be collected autonomously based on the number of failures. Both CCG and TR play a crucial role in reducing the number of failures to zero, which allowed us to collect the necessary amount of data seamlessly.}
  \label{fig:robust}
\end{figure}

We conducted an ablation study on CCG and TR where we ran self-supervised data collection for a total of 128 episodes and recorded the number of failures.
We visualize the results in \Cref{fig:robust}.
The naive approach without CCG and without TR failed a total of 15 times, which roughly translates to an average of nine episodes collected autonomously before a human intervention is needed.
The number of failures dropped to three with the addition of CCG, which raises the average to 43 episodes collected before each intervention. 
Finally, with both CCG and TR, we were able to successfully collect 128 episodes without any human intervention.

\subsection{Noise Augmentation Ablation}
\label{subsec:noise}

\begin{table}[b]
	\vspace{-2mm}
	\centering
	\caption{Success rates for a model trained on a dataset collected with and without noise augmentation and using two different policy action representations. The noise-augmented data collection procedure improves the performance of both deterministic and Gaussian mixture policies.}
	\renewcommand{\arraystretch}{1.1}
	\begin{tabular}{lccc}
	\toprule
	\multicolumn{1}{l}{\textbf{\thead{Action \\ Representation}}} & \textbf{\thead{Trained \\ with Noise \\ Aug. Data}} & \multicolumn{1}{c}{\textbf{\thead{Avg. \\ Success Rate}} ($\uparrow$)} & \multicolumn{1}{c}{\textbf{\thead{Total \\ Successes}} ($\uparrow$)}\\ \midrule
	\multirow{2}{*}{Deterministic}     & No & 71.67(5.93) & 57/80                   \\ 
	    & Yes & 81.11(5.50) & 56/70                 \\ 
	\multirow{2}{*}{Gaussian Mixture}    & No & 72.22(5.67) & 50/70         \\ 
	    & \textbf{Yes} & \textbf{87.78(3.93)} & \textbf{62/70}         \\ 
%   No clearance pose sampling & - & - \\
% 	With tactile sensing    & - & -         \\ 
	\bottomrule
	\end{tabular}	
	\label{tab:policy}
\end{table}
In \Cref{subsec:noiseaugment}, we introduced a noise-augmented data collection variant of PvP.
We investigated its contribution to the success rate of the policy.
We collected two different datasets each containing 128 demonstrations, with and without added noise augmentation.
We tested the effect of noise-augmented data on policies trained using two different action representations: a deterministic policy and a Gaussian mixture policy with five modes. 
We present the results in \Cref{tab:policy}.
For each row, we averaged the success rate over three models trained from scratch with three different seeds.
We used 20 or 30 rollouts to calculate the success rate of each model.
The numbers in the brackets denote variation of one standard deviation.
We also report the total successes summed over the three models.
For both action representations, we found that training on noise-augmented data improved the performance of the final policy significantly. 
The deterministic policy had an improvement of approximately 10\% while the mixture policy had a larger improvement of 15\%. 

We hypothesize that adding noise further accentuates the multimodal nature of the data (i.e., multiple valid actions are available for roughly the same observation) and that a mixture representation for the actions allows the policy to better capture this by not having to average over multiple possible expert actions \cite{bishop1994mixture}. 
We also qualitatively observed that policies trained with noise-augmented datasets were better at re-adjusting the object with respect to the goal. 

\subsection{Comparison to Kinesthetic Teaching}
\label{subsec:comparison}

\begin{figure}[t!]
    \includegraphics[height=5cm,keepaspectratio]{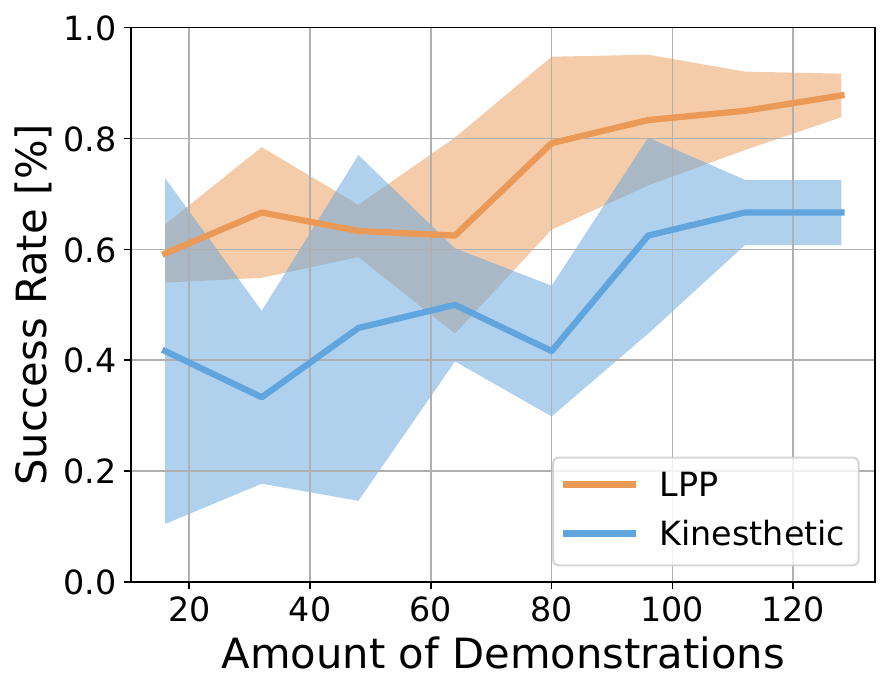}
    \centering
    \caption{Comparison of success rates of policies trained using a dataset collected by PvP and a traditional kinesthetic teaching approach. We tested models trained on an increasing amount of demonstration data. Policies trained on data collected by PvP outperform those trained on data collected from kinesthetic teaching for almost all dataset sizes. The shaded region consists of one standard deviation.} 
	\label{fig:comparison}
	\vspace{-2mm}
\end{figure}

We compared the performance of policies trained using demonstration data from PvP and from kinesthetic teaching. 
In particular, we investigated whether the quality of the demonstrations differ. 
To compare, we trained two separate policies, both sharing the architecture outlined in \Cref{subsec:policy}, on two separate datasets: one collected with PvP and another collected with kinesthetic teaching \cite{billardLearningHumans2016}. 
We used a Gaussian mixture model as the action representation of both policies.
For PvP, we used the noise-augmented variant since it was the best-performing model, as shown in \Cref{subsec:noise}.

In \Cref{fig:comparison}, we visualize the success rates of both models trained with varying numbers of demonstrations.
As done previously, we averaged the performance over three models with different seeds for each datapoint. 
We tested each model using eight rollouts. 
The models trained with data from PvP, collected without any human involvement, significantly outperform the models trained using kinesthetic data by about 20\% for almost all numbers of demonstrations.
We observed that policies trained with kinesthetic teaching often misplaced the plate (e.g., by not having the right place angle or not resting the plate in a stable manner with respect to the environment contact points) and struggled to open the gripper at the correct times.
The latter of the two error modes did not appear in policies trained with PvP.

We hypothesize that the kinesthetic policy performed more poorly than the PvP policy because the overall quality of the human demonstrations was lower than the programmatic or machine-generated demonstrations produced by PvP.
Previous work has studied the difficulties and challenges of training policies using human demonstrations when compared to programmatic or machine-generated demonstrations \cite{KostrikovADLT19, orsini2021matters, Mandlekar2021WhatMI}. 
In particular, human demonstrations can be sub-optimal due to mistakes (e.g., large variations in trajectory length, unwanted movements, and remaining idle for some time to think).  
The authors of \cite{Mandlekar2021WhatMI} note that the relative average trajectory length is a good proxy for the quality of the dataset.
The average number of time steps for all demonstrations was 29.41 (2.22) for the PvP dataset and 41.66 (5.69) for the kinesthetic dataset.
The numbers in the brackets denote variation of one standard deviation. 
Kinesthetic demonstrations were on average longer and had greater variance in length. 
In addition, PvP benefits from systematic noise-augmentation, as covered in \Cref{subsec:noiseaugment}, which a human demonstrator cannot emulate consistently.

\section{Limitations and Future Work} \label{sec:limitation}
PvP relies on effective grasping (i.e., accurate grasp detections and stable grasps during retrieval).
This makes this method susceptible to the typical challenges involved in grasping, namely dealing with noisy point cloud measurements, unreliable object segmentation, and unmodelled object dynamics (e.g. object slip). 
The quantity, quality, and diversity of the grasp poses found by the planner are partly determined by the viewpoint of the camera.
In this work, we empirically selected a viewpoint that is adequate for our use case.
It would be interesting to allow the manipulator to explore and find better grasp poses.
Furthermore, while tactile sensing is used to improve the robustness of data collection, it could also be used as an extra modality for the policy.
We have introduced a framework for data collection and tested it with two real-world tasks.
We would like to test PvP on a larger set of tasks for a more exhaustive evaluation.
Lastly, PvP is a language-driven data collection method, and it would be natural to integrate it into a high-level language based planner \cite{gdm2024autort}.

\section{Conclusion} \label{sec:con}

In this paper, we presented PvP, a self-supervised data collection method to collect expert demonstrations for placement tasks. 
PvP leverages recent advancements in grasp planners, tactile sensing, compliant control, and the symmetry between picking and placing to autonomously collect placing demonstrations.
PvP is able to robustly collect a large amount of data consistently without human intervention by making use of compliant manipulator control and regrasping based on tactile sensing.
We showed that the quality of the demonstrations produced by PvP surpasses that of human demonstrations collected by kinesthetic teaching.
PvP provides a promising and pragmatic direction to collect real world robotic placing demonstration data with minimal human labour. 




% \section*{ACKNOWLEDGMENT}

% The preferred spelling of the word ÒacknowledgmentÓ in America is without an ÒeÓ after the ÒgÓ. Avoid the stilted expression, ÒOne of us (R. B. G.) thanks . . .Ó  Instead, try ÒR. B. G. thanksÓ. Put sponsor acknowledgments in the unnumbered footnote on the first page.

\balance
\bibliographystyle{ieeetr}
\bibliography{refs}

\end{document}